\documentclass[11pt,a4paper]{article}
\usepackage[hyperref]{acl2021}
\usepackage{times}
\usepackage{latexsym}

\usepackage{scalerel}

\usepackage{microtype}
\usepackage{multirow}
\usepackage{adjustbox}
\usepackage{graphicx}
\usepackage{subcaption}
\usepackage{diagbox}
\usepackage{makecell}
\usepackage{booktabs}
\usepackage{verbatim}
\usepackage{caption} 
\aclfinalcopy

\title{{CONDA}: a {CON}textual Dual-Annotated dataset \\for in-game toxicity understanding and detection}

\author{
Henry Weld \qquad
Guanghao Huang \qquad
Jean Lee \qquad
Tongshu Zhang \qquad
Kunze Wang \\ \qquad
\textbf{Xinghong Guo} \qquad
\textbf{Siqu Long} \qquad
\textbf{Josiah Poon} \qquad
\textbf{Soyeon Caren Han\thanks{\, Corresponding author (caren.han@sydney.edu.au)}}
\\
School of Computer Science, The University of Sydney, NSW, Australia \\
\texttt{\{hgua3108,tzha6458,kwan4418,xguo2796,slon6753\}@uni.sydney.edu.au} \\
\texttt{\{henry.weld,jean.lee,josiah.poon,caren.han\}@sydney.edu.au} \\
  }

\date{}

\begin{document}
\maketitle
\begin{abstract}
Traditional toxicity detection models have focused on the single utterance level without deeper understanding of context. We introduce CONDA, a new dataset for in-game toxic language detection enabling joint intent classification and slot filling analysis, which is the core task of Natural Language Understanding (NLU). The dataset consists of 45K utterances from 12K conversations from the chat logs of 1.9K completed \textit{Dota 2} matches. We propose a robust dual semantic-level toxicity framework, which handles utterance and token-level patterns, and rich contextual chatting history. Accompanying the dataset is a thorough in-game toxicity analysis, which provides comprehensive understanding of context at utterance, token, and dual levels. Inspired by NLU, we also apply its metrics to the toxicity detection tasks for assessing toxicity and game-specific aspects. We evaluate strong NLU models on CONDA, providing fine-grained results for different intent classes and slot classes. Furthermore, we examine the coverage of toxicity nature in our dataset by comparing it with other toxicity datasets.\footnote{The dataset and lexicons are available at https:// github.com/usydnlp.}

\end{abstract}

\section{Introduction}
As the popularity of multi-player online games has grown, the phenomenon of in-game toxic behavior has taken root within them. Toxic behavior is strongly present in recent online games and is problematic to the gaming industry \cite{adinolf2018toxic}. For instance, 74\% of US players of such games report harassment with 65\% experiencing severe harassment. \cite{adl2019free}.


In the past few years, Natural Language Processing (NLP) researchers have proposed several online game/community toxicity analysis frameworks \cite{KwakBH15, Murnion18, wang2020detect} and datasets \cite{MartensSIK15, stoopetal19}. However, existing datasets (1) focus only on the single utterance level without deeper understanding of context in the whole conversation/chat, and (2) do not explicitly use semantic clues from the words within the utterance.

\begin{figure}[t]
         \centering
         \includegraphics[width=\columnwidth]{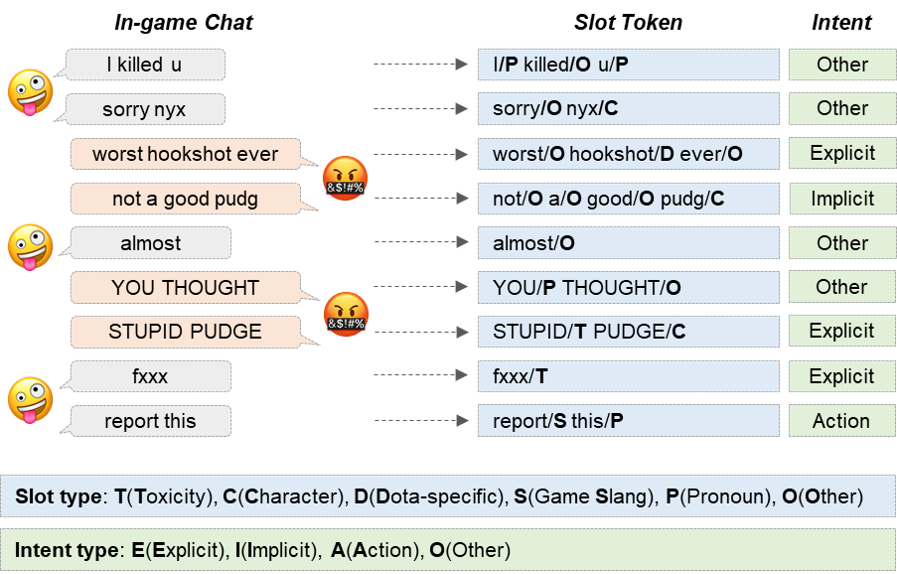}
    \caption{An example intent/slot annotation from the CONDA (CONtextual Dual-Annotated) dataset.}
    \label{fig:annotation}
\end{figure}

The chat in online games and communities is similar in nature to spoken language, an area studied by Natural Language Understanding (NLU). NLU research aims to best represent human communication by extracting semantic structure in the form of intent and slot analysis. Intent detection is the classification of the desired outcome of an utterance (or sentence), and slot filling is the labeling of each token (or word) in the utterance with the type of semantic information it carries. In recent literature, these two tasks are trained jointly to capture synergies between them, and these jointly trained models give better results \cite{ZhangLDFY19}. Furthermore, researchers have made available joint task datasets that contain the context of a multi-turn conversation \cite{BudzianowskiWTC18, SchusterGSL19}

Inspired by this NLU research progress, we propose CONDA, an in-game toxicity detection dataset, with a robust dual-level annotation which enables intent detection and slot filling. Our dataset consists of 45k utterances from the chat logs of 1.9k \textit{Dota 2} matches, labeled with 4 intent classes and 6 slot classes to address toxicity and the game-specific vocabulary. Figure \ref{fig:annotation} illustrates an example of CONDA including raw data (in-game chat) and processed data with slot and intent labels. In order to enable the dual semantic-level framework, we conduct lexicon-based automation for token-level data and human annotation for utterance-level data. 

We investigate the CONDA dataset through an in-depth analysis. The large portion of game-specific classes in the dual levels enables the dataset to be more sophisticated in detecting toxicity in games. The combination of each intent with each slot class shows that dual annotation can help determine toxicity from gamer slang when used in both toxic and non-toxic situations. We also find more toxic utterances appear pre-game and post-game rather than during the games, especially peaking post-game due to the chat for post-victory celebration and recrimination.

We provide five strong baseline NLU models and compare the toxicity detection performance over our dataset. For evaluation, we apply four NLU metrics to assess performance in toxicity and game specific aspects. Results vary across models, indicating a challenge for improvement. Furthermore, we perform a transfer learning experiment with existing toxicity datasets. We find that the nature of toxicity in our dataset can generalize to other proposed taxonomies, including hatefulness, sexism and racism. Beyond this commonality, our experiment illustrates that CONDA is distinguished from other toxicity datasets due to game-specific characteristics. This paper then makes the following contributions:
\begin{itemize}
\setlength\itemsep{0em}
\item To the best of our knowledge, this is the first attempt to build a toxicity detection dataset with joint Natural Language Understanding aspects of intent classification and slot filling;

\item We propose a robust dual semantic-level toxicity framework, which handles utterance and token-level patterns with rich in-game chatting history;

\item We formalise NLU metrics for toxicity detection, evaluate strong NLU models on our dataset, and further conduct transfer learning experiments with other toxicity datasets. 

\end{itemize}

\begin{table*}[t!]
\centering
\small
\setlength{\tabcolsep}{1.6mm}{
\begin{tabular}{llllc}
\toprule
\multicolumn{1}{l}{\textbf{Dataset}} &
  \multicolumn{1}{l}{\textbf{Approach}} &
  \multicolumn{1}{l}{\textbf{Domain}} &
  \multicolumn{1}{l}{\textbf{Labels}} &
  \multicolumn{1}{l}{\textbf{Conv.}} \\ \midrule
\cite{MartensSIK15} &
  utterance-level &
  Game (Dota 2) &
  toxic, non-toxic &
  N \\ 
\cite{WaseemH16} &
  utterance-level &
  Twitter &
  racist, sexist, normal &
  N \\ 
\cite{NobataTTMC16} &
  utterance-level &
  Yahoo News &
  clean, hate, derogatory, profanity &
  N \\
\cite{DavidsonWMW17} &
  utterance-level &
  Twitter &
  hateful, offensive, neither &
  N \\
\cite{GaoH17} &
  utterance-level &
  Fox News &
  hate, non-hate &
  N \\
\cite{ElSheriefNNVB18} &
  utterance-level &
  Twitter &
  hate, non-hate / hate instigator, hate target &
  N \\ 
\cite{FountaDCLBSVSK18} &
  utterance-level &
  Twitter &
  \begin{tabular}[c]{@{}l@{}}offensive, abusive, hateful speech, \\ aggressive, cyberbullying, spam, normal\end{tabular} &
  N \\ 
\cite{Mizil18} &
  utterance-level &
  Wikipedia &
  toxic, non-toxic &
  Y \\ 
\cite{stoopetal19} &
  utterance-level &
  Game (LoL) &
  toxic, non-toxic &
  Y \\ 
\cite{QianBLBW19} &
  utterance-level &
  Gab \& Reddit &
  hate, non-hate &
  Y \\ 
\cite{PavlopoulosSDTA20} &
  utterance-level &
  Wikipedia &
  toxic, non-toxic &
  Y \\ 
\textbf{CONDA (our dataset)} &
  \textbf{\begin{tabular}[l]{@{}l@{}}dual-level \\ (utterance and token)\end{tabular}} &
  Game (Dota 2) &
  \begin{tabular}[l]{@{}l@{}}- utterance level (intent): explicit toxicity, \\ implicit toxicity, action, others\\ - token level (slot): toxicity, character, \\ dota-specific, slang, pronoun, other\end{tabular} &
  \textbf{Y} \\ \bottomrule
\end{tabular}%
}
\setlength{\belowcaptionskip}{-10pt}
\caption{Comparison of CONDA with other toxicity datasets (Conv.: Conversation).}
\label{tab:dataset_comparisons}
\end{table*}

\section{Related Work}
\paragraph{Toxicity Datasets in Online Games}
In multiplayer online games, prior research focused on analysis of anti-social or disruptive behavior, so-called toxic behavior \cite{BlackburnK14, NetoB18} including cyberbullying \cite{KwakBH15} and griefing \cite{Murnion18}. Although these terms contain similar elements, a single definition of toxic behavior is yet to emerge. Some studies have conducted data annotation using pre-defined lexicon categories \cite{MartensSIK15} or toxic player information \cite{stoopetal19}. These annotation methods are not robust enough to handle unlabelled toxicity words or unreported toxic players. 

\paragraph{Toxicity Datasets in Online Community}
An extensive body of work has focused on datasets to detect toxicity including hate speech \cite{WaseemH16, DavidsonWMW17, ElSheriefNNVB18} and abusive language \cite{NobataTTMC16, FountaDCLBSVSK18}. However, the majority of toxicity datasets do not consider the context of a conversation, instead simply analysing a single utterance. Even if a model uses contextual information \cite{GaoH17}, it is limited to meta-information (e.g. news title or user name) which is not sufficient to understand a conversation. In our research, context is defined as linguistic contextual information, particularly previous single or multiple utterances. Along similar lines, recent studies have focused on conversation aiming to discover warning signals \cite{Mizil18}, to generate intervention responses \cite{QianBLBW19}, or to measure the importance of context \cite{PavlopoulosSDTA20}. Existing toxicity datasets mainly focus on annotating at utterance-level, whereas ours conducts a dual-level annotation at utterance and token-level, while also providing a conversation history (see Table \ref{tab:dataset_comparisons}). These extra features are what distinguish CONDA.

\paragraph{NLU Datasets and Models}
In-game chat has similar characteristics to multi-turn dialogue in NLU. The approaches used in multi-turn dialogue analysis have not yet been observed in toxicity datasets. In NLU, generally, intent classification (IC) is treated as a semantic utterance classification task and slot filling (SF) is treated as a sequential token labelling task \cite{ZhangW16a}. By conducting a joint model for the two tasks, a synergistic effect can be achieved \cite{ZhangLDFY19}. To build multi-turn dialogue datasets, most studies have recruited workers via crowd-sourcing to collect task-oriented dialogues across different domains (e.g. in-car assistant \cite{EricKCM17}, navigation and events \cite{GuptaSMKL18}, multi-domains \cite{BudzianowskiWTC18}, personal notifications \cite{SchusterGSL19}). Recently, deep learning models have also been extensively studied in order to capture the contextual signals from multiple sequential inputs. (e.g. BiLSTM with attention \cite{WangHFZT19}, GRU with self-attention and context-fusion \cite{GuptaZLD19}. The models listed all show an increase in semantic detection performance when the context is included in the analysis.

\section{CONDA}

\subsection{Data Collection}\label{data_collection}
Our annotated dataset, CONDA, is based on the \textit{Defense of the Ancients 2 (Dota 2)} data dump available at Kaggle\footnote{https://www.kaggle.com/devinanzelmo/dota-2-matches}. \textit{Dota 2} is a multiplayer online game where teams of five players attempt to destroy their opponents' ancient structure. The raw data is compiled from game matches including players, duration, match outcomes, and complete chat logs. In order to curate data, we select 50,000 utterances in complete chat logs from 1,921 matches. 

\subsection{Data Processing}\label{data_processing}
Our data processing is designed to enable dual annotation, making utterance-level data suitable for human annotators and generating token-level data for lexicon-based automation. The main processes are creation of conversations, restructuring utterances while keeping original context, and generation of tokens. 

We generate conversations to give human annotators a context of previous utterances when labelling the current utterance. We identify the beginning of a conversation as the first utterance in the match, or an utterance that occurs greater than 60 seconds after the previous utterance in the match. While the raw data is largely in English, other languages appear occasionally including Russian, Chinese, Spanish, etc. We exclude conversations with chat in non-English.

For the utterance-level data, we maintain the original form such as punctuation and case in order to keep context. In addition, we merge consecutive utterances by a single user within a conversation. These are combined into one utterance with a special token, [SEPA], added to denote the separation point (e.g. \textit{`easiest [SEPA] game [SEPA] of my life'}). For the token-level data, we use contraction restoration (e.g. \textit{`I’m' -$>$ `I am'}), whitespace tokenise each utterance, retain emoticons, but remove punctuation. This token-level processing is used for lexicon-based automated slot annotation. 

Our final CONDA dataset (Table \ref{tab:CONDA and subset}) consists of 44,869 utterances and 1,921 matches. We further create a subset, equivalent to about 10\% of the full dataset, for a preliminary round of utterance-level annotation.


\begin{table}[ht]
\centering
\small
\setlength{\tabcolsep}{8mm}{
\begin{tabular}{ll}
  \toprule
  \textbf{Dataset Feature} & \textbf{CONDA} \\
   \midrule
  Matches & 1,921 \\ 
  Conversations & 12,152 \\ 
  Utterances & 44,869 \\ 
  Avg. utterances per match & 23.3 \\ 
  \bottomrule
\end{tabular}}
\setlength{\belowcaptionskip}{-10pt}
\caption{CONDA statistics.}
\label{tab:CONDA and subset}
\end{table}

\subsection{Annotation}
\paragraph{Dual Aspects}
Inspired by NLU, we provide a dual-level annotation approach to detect toxicity, which often relies on context. This allows one to find toxic intent even though an utterance does not contain any toxic words, or to determine non-toxic intent even if an utterance has toxic words. For example, Figure \ref{fig:annotation} shows an utterance \textit{“not a good pudg”}, which does not contain any toxic words. However, considering the previous utterance of \textit{“worst hookshot ever”}, we can identify hidden or implicit toxicity. As an example of the other way around, an utterance of \textit{“happy fuck you day”} contains a toxic word but it is used for cheering after saying \textit{``gg" (good game)}. 

\paragraph{Token-level Slot Annotation}
With the processed token-level data, an automated slot labelling is performed. Initially, we create six distinct slot labels: T (\textbf{T}oxicity), C (\textbf{C}haracter), D (\textbf{D}ota-specific), S (game \textbf{S}lang), P (\textbf{P}ronoun) and O (\textbf{O}ther). To construct the \textbf{T} lexicon, we combine several toxicity lexicons (see Section \ref{sec:ethic} Ethics) and remove overlaps. We also use the supplemental data sourced by \citet{MartensSIK15} for the game-related lexicons (\textbf{C, D} and \textbf{S}) and carefully modify it. The \textbf{P} lexicon (e.g. \textit{`u', `ur'}) is constructed by this research because in-game chat is extremely abbreviated.  Then, we perform lexicon-based automation by exact matching each lower-cased token against the lexicons. Anything not matching a lexicon is labelled \textbf{O}. We contrast this with typical NLU slot labelling where a semantic concept can stretch over a span of words. In comparison to other toxicity datasets, our lexicon-based slot labelling enables deeper understanding of game context. 


\paragraph{Utterance-level Intent Annotation}
Given tokens with slot labelling and utterance-level data, we perform a test run on the subset of utterances using six annotators. Four annotators are game players and two are non-game players. This preliminary round is for fine-tuning annotation policy and analysing annotator agreement to inform final annotation for the full dataset. The annotators manually classified the utterances into four labels:  E (\textbf{E}xplicit toxicity), I (\textbf{I}mplicit toxicity), A (\textbf{A}ction) and O (\textbf{O}ther). The label details are explained in the annotator instructions.



\paragraph{Annotator Instructions}
Each annotator was required to consider the earlier conversation, particularly, to detect implicit toxic behavior or to identify non-toxic behavior in the utterance having toxic-labelled tokens. The annotators worked independently of one another. The guidelines for human annotators were as follows:
 
\textbf{E}xplicit toxicity: Typically contains toxic word(s). The intent is to insult or humiliate others, or to make others want to leave the conversation or quit the game. There is no need to consider the context (e.g. \textit{`fuck off'}). May include one or more of the following aspects: 
\begin{itemize}
\setlength\itemsep{0em}
\item Strong toxicity - blatant insulting or disrespecting others is obviously seen in the text, normally with severely toxic wording;
\item Normal toxicity - impolite, rudely worded and unreasonable comment that insults or humiliates others; 
\item Cursing others with the intent to insult or humiliate them (e.g. \textit{`noob'}\footnote{“Noob” is a slang term for a newcomer, commonly used to insult someone inexperienced in games.});
\item Sexual wording or talk about sex-related behavior; 
\item Use of negative or hateful words to describe others (e.g. \textit{`useless'}); 
\item Racist language that is targeted at insulting others (e.g. \textit{`Peruvians', `fucking russians'}); 
\item Inflammatory language, insulting others and trying to start a conversational fight. 
\end{itemize}

\textbf{I}mplicit toxicity: Hidden toxicity that normally cannot be seen from the text itself. The text might be factual or even positive (e.g. sarcasm). However, based on the utterance or conversation context, the intent of insulting or humiliating others can be inferred. Typically contains no toxic word (e.g. \textit{`u are poor dude'}).

\textbf{A}ction: Doesn't belong to I or E, but contains an action such as report, commend, pause, stop, or exit game. 

\textbf{O}ther: Doesn't belong to I or E or A. May or may not contain toxic words. Includes curses, self-deprecation or any other emotional expression that is NOT targeted at others (e.g. \textit{`kill the fucking helicopter'}).

\paragraph{Findings in Annotation}
The preliminary round was useful for enabling discussion around annotation. For example, we decided that the token “ez”\footnote{“Ez” is an abbreviation for easy. It is often used to irritate other players in games, indicating “You are just way too easy”.} or its variations are an implied slur against the opposition's quality and would generally be part of an \textbf{I} label utterance. Similarly, “g” is often a contraction for “go” and would be part of an \textbf{A} label utterance. 
Overall, we observed that the agreement measure for utterance classification was higher for gamer annotators only (Fleiss' kappa = 0.785) versus the whole group (Fleiss' kappa = 0.755). The lower inter-rater agreement in the whole group is because non-gamer annotators have low understanding of the game context and domain-specific language. Therefore, annotation of the whole dataset was performed by gamers only. Based on our annotation guidelines, they collectively manually annotated the utterances for the full dataset.  


\section{Dataset Analysis}
The CONDA dataset consists of 9 columns - match ID, conversation ID, player ID, player slot, chat time, utterance, slot tokens (cleaned tokens with slot labelling), intent class, and slot classes. For example, the utterance \textit{“gg wp”} for \textit{“good game well played”} is shown in the slot tokens column as \textit{“gg (S), wp (S)”}. Each column is further explained in AppendiX \ref{sec:conda_column}. 

\begin{table}[!ht]
\small
\centering
\begin{minipage}[t]{0.44\linewidth}\centering
    \setlength{\tabcolsep}{1mm}{
        \begin{tabular}{lcc}
        \toprule
          \textbf{Intent} & \textbf{\%}  & \textbf{Mean L.} \\
          \midrule
          E (\textbf{E}xplicit) & 13.3   & 6.14  \\
          I (\textbf{I}mplicit) & 6.4   & 4.16  \\
          A (\textbf{A}ction) & 6.4  & 4.40  \\
          O (\textbf{O}ther) & 73.9   & 3.18 \\
          \midrule
           \textbf{Total} & \textbf{100.0} &  \textbf{3.71}  \\
          \bottomrule
        \end{tabular}
        }
\end{minipage}\hfill%
\begin{minipage}[t]{0.44\linewidth}\centering
        \setlength{\tabcolsep}{1mm}{
        \begin{tabular}{lc}
        \toprule
          \textbf{Slot} & \textbf{\%}  \\
          \midrule
          T (\textbf{T}oxicity) & 4.9   \\
          C (\textbf{C}haracter) & 5.4   \\
          D (\textbf{D}ota-specific) & 1.4  \\
          S (Game \textbf{S}lang) & 11.2 \\
          P (\textbf{P}ronoun) & 13.5   \\
          O (\textbf{O}ther) & 63.6 \\
          \midrule
        \textbf{Total} & \textbf{100.0}   \\
          \bottomrule
        \end{tabular}
        }
\end{minipage}
        \caption{Intent labelling statistics (left) and slot labelling statistics (right). $\%$ is proportion of the dataset. Mean L. is mean number of tokens after cleaning.}
        \label{tab:intents_with_punc_SEPA_removed}
\end{table}

\paragraph{Dual Annotation Proportion} \label{sec:datasetstats}
Table \ref{tab:intents_with_punc_SEPA_removed} gives the proportional breakdown of the CONDA dataset by intent and slot labels. Together the toxic utterance classes make up 19.7\% of the data, emphasizing their prevalence in game chat. The proportion of the I (6.4\%) and A (6.4\%) intent classes, together 12.7\% of all utterances, shows the more granular non-binary class structure captures an aspect of online games. Additionally, the average length of utterances of the E class (6.14) is greater than for each other class, indicating players strongly emphasize emotional frustration. In the proportion of slot labels, we can see the S (11.2\%) class is more than double the C (5.4\%) class and 8 times the D (1.4\%) class, indicating general gamer slang is used for communication more than terms specific to the game being played. Overall, the large portion of these game-specific classes enables the dataset to be more sophisticated in detecting toxicity in games. 


\begin{figure} [t]    

     \begin{subfigure}{0.49\linewidth}
         \includegraphics[width=\linewidth]{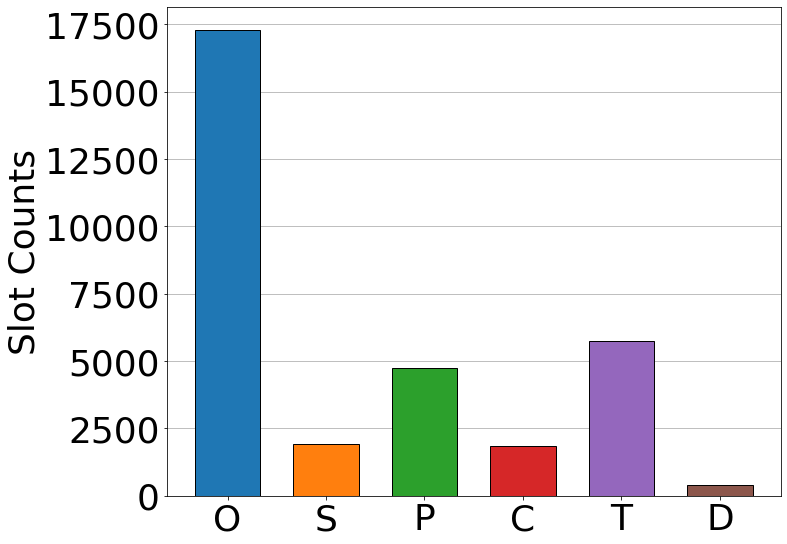}
         \vspace*{-7mm}
         \caption{Class 'Explicit'}
         \label{fig:Eslot}
     \end{subfigure}
     \begin{subfigure}{0.49\linewidth}
         \includegraphics[width=\linewidth]{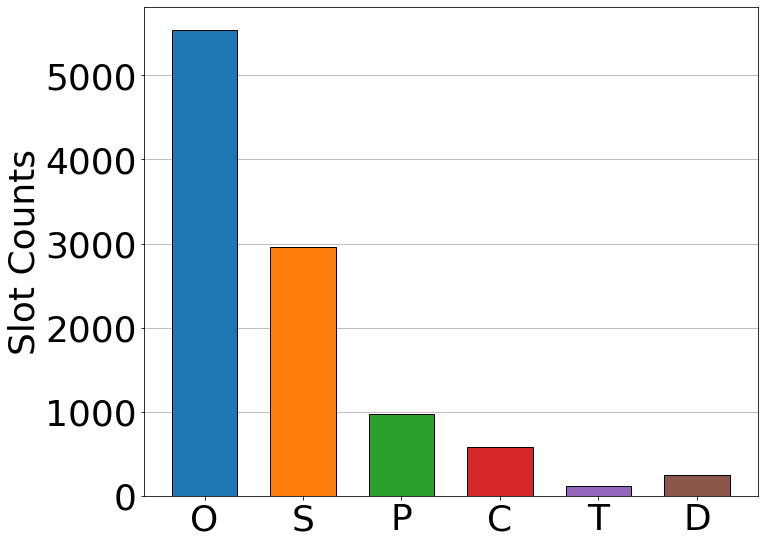}
         \vspace*{-7mm}
         \caption{Class 'Implicit'}
         \label{fig:Islot}
     \end{subfigure}
    \begin{subfigure}{0.49\linewidth}
         \includegraphics[width=\linewidth]{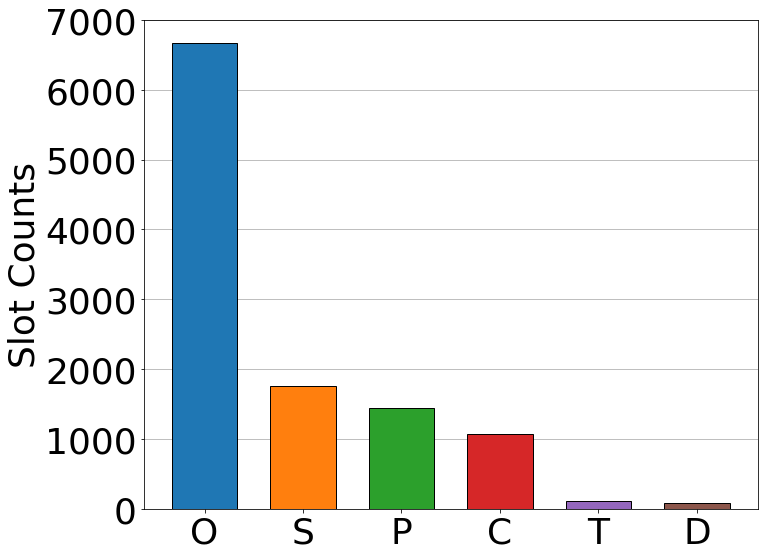}
         \vspace*{-7mm}
         \caption{Class 'Action'}
         \label{fig:Aslot}
     \end{subfigure}
     \begin{subfigure}{0.49\linewidth}
         \includegraphics[width=\linewidth]{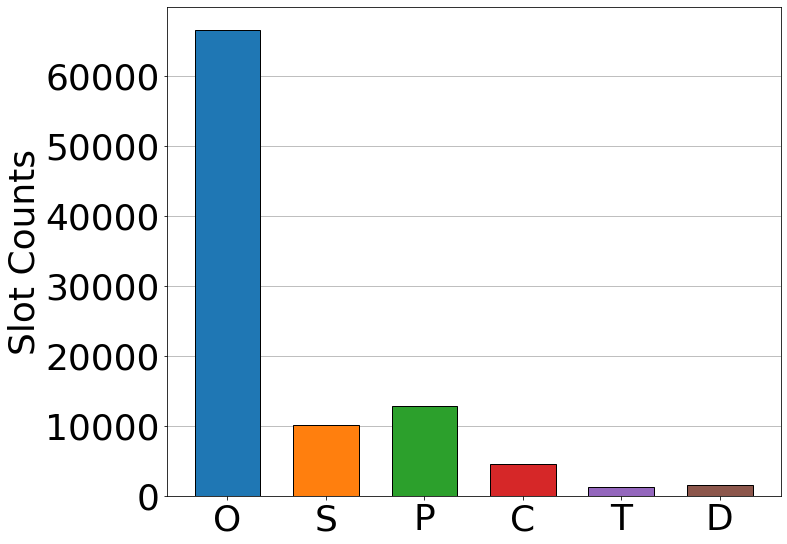}
         \vspace*{-7mm}
         \caption{Class 'Other'}
         \label{fig:Oslot}
     \end{subfigure}
     \setlength{\belowcaptionskip}{-10pt}
     \caption{Slot class distributions for each intent class.}
     \label{fig:slot_by_intent}
     
\end{figure}

\begin{table}[t]
\centering
\scriptsize
\begin{minipage}[t]{0.5\linewidth}\centering
\centering
    \scalebox{0.85}{
        \begin{tabular}{lcc}
        \toprule
          \textbf{Rank} & \textbf{S} & \textbf{T}\\
          \midrule
          1 & gg (239) & noob (878)  \\
          2 & report (237) & fuck (807) \\
          3 & ez (191) & fucking (593) \\
          4 & mid (169) & shit (546) \\
          5 & go (114) & idiot (222) \\
          \bottomrule \\

          \multicolumn{3}{c}{\footnotesize (a) Class "Explicit"} \\
          & & 
        \end{tabular}
        }

\end{minipage}\hfill%
\begin{minipage}[t]{0.5\linewidth}\centering
    \scalebox{0.85}{
        \begin{tabular}{lcc}
        \toprule
          \textbf{Rank} & \textbf{S} & \textbf{T}  \\
          \midrule
          1 & ez (1,932) & wtf (32)\\
          2 & mid (287) & fucking (11) \\
          3 & gg (169) & dead (9) \\
          4 & report (47) & hook (6) \\
          5 & go (38) & fuck (5) \\
          \bottomrule \\

          \multicolumn{3}{c}{\footnotesize (b) Class "Implicit"} \\
          & & 
        \end{tabular}
        }
\end{minipage}\hfill%
\begin{minipage}[t]{0.5\linewidth}\centering
    \scalebox{0.85}{
        \begin{tabular}{lcc}
        \toprule
          \textbf{Rank} & \textbf{S} & \textbf{T} \\
          \midrule
          1 & report (992) & wtf (16) \\
          2 & afk (184) & fucking (11) \\
          3 & gg (134) & abuse (10) \\
          4 & go (57) & noob (8) \\
          5 & wp (37) & shit (4) \\
          \bottomrule \\
          \multicolumn{3}{c}{\footnotesize (c) Class "Action"} \\
        \end{tabular}
        }
\end{minipage}\hfill%
\begin{minipage}[t]{0.5\linewidth}\centering
    \scalebox{0.85}{
        \begin{tabular}{lcc}
        \toprule
          \textbf{Rank} & \textbf{S} & \textbf{T}  \\
          \midrule
          1 & gg (3,735) & wtf (331)\\
          2 & wp (1,115) & dead (89) \\
          3 & ggwp (776) & fucking (84) \\
          4 & mid (413) & hook (69) \\
          5 & go (383) & shit (39) \\
          \bottomrule \\
          \multicolumn{3}{c}{\footnotesize (d) Class "Other"} \\

        \end{tabular}
        }
\end{minipage}
\setlength{\belowcaptionskip}{-10pt}
\caption{Top 5 keywords in the S (game Slang) and T (Toxicity) slot classes, for each intent class. The number in brackets is the token count in that combination of classes.}
\label{tab:keyword}
\end{table}

\paragraph{Dual Annotation Distribution}
To understand the effect of dual annotation on the toxicity context, we look at the distribution of the slot labels within each intent class. As seen in Figure \ref{fig:Islot}, the I intent class shows the highest proportion of the S slot class among non-O classes. Similarly, Figure \ref{fig:Aslot} shows a relatively high proportion of the S slot class in the A intent class. This suggests that the combination of S slot and intent classes provides useful information because slang performs the function of carrying game-specific context. As a result, we focus more on T and S slot classes joined with other intent classes in order to investigate toxicity natures in games carried out from dual annotation.  

\paragraph{Keywords in Dual Annotation}
Table \ref{tab:keyword} shows the top 5 keywords by frequency from the T and S slot classes, for each intent class. In the combinations of S class, we observe the prominent position of “ez” in the E and I intent class. This indicates dual annotation captures toxicity from the slang largely used in games. In addition, “gg” appears in all combinations because it may have some toxicity attached via sarcasm. As dual annotation uses the conversational history, it is able to classify the same utterance in different intents. 



\begin{figure}[t]    
     \begin{subfigure}{0.49\linewidth}
         \includegraphics[width=\linewidth]{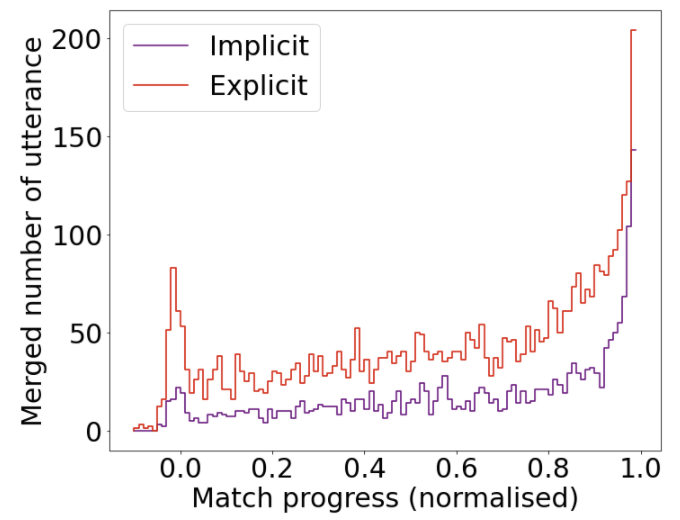}
         \vspace*{-5mm}
         \caption{Class “Explicit, Implicit”}
         \label{fig:duration_EI}
     \end{subfigure}
     \begin{subfigure}{0.49\linewidth}
         \includegraphics[width=\linewidth]{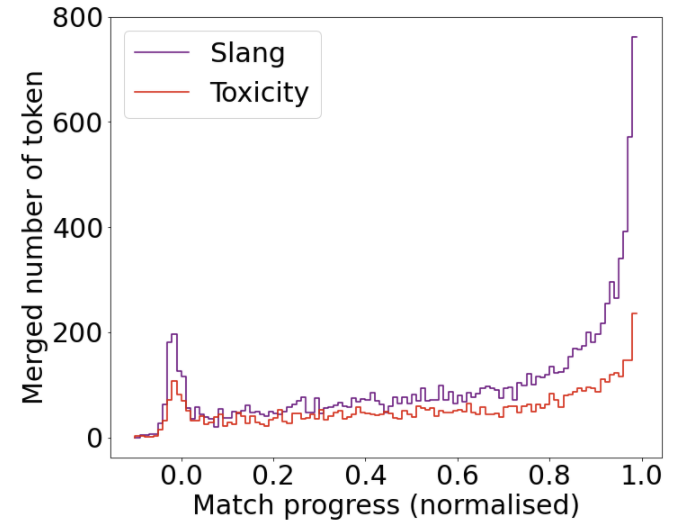}
         \vspace*{-5mm}
         \caption{Class “Toxicity, Slang”}
         \label{fig:duration_TS}
     \end{subfigure}
     \caption{In-game chat histogram for intent (E,I) and slot (T,S) classes. Match progress is bucketed position within a match whose duration is normalised in [0,1], with $<$0 indicating pregame chat. Merged number of utterance/token is the count of all utterances/tokens from all matches in that match progress bin.}
     \label{fig:duration_matches}
\end{figure}

\paragraph{Toxicity Analysis Over Time}
We further analyse in-game chat over time associated with the E, I intent classes and the T, S slot classes. As shown in Figure \ref{fig:duration_EI}, more toxic utterances appear pre-game and post-game rather than during the games. In pre-game, sometimes players are upset due to the selected hero characters if their desired hero is taken by others, or are stressed by planning game strategies in a limited time. Toxic utterance frequency gradually rises towards the end, and peaks in post-game due to chat for post-victory celebration and recrimination. Interestingly, Figure \ref{fig:duration_TS} displays a similar pattern. Particularly, tokens for the S slot class increase sharply to the end, indicating significant amounts of slang are used to celebrate wins or humiliate defeated opponents. 

\begin{figure}[t]
         \centering
         \includegraphics[width=0.8\columnwidth]{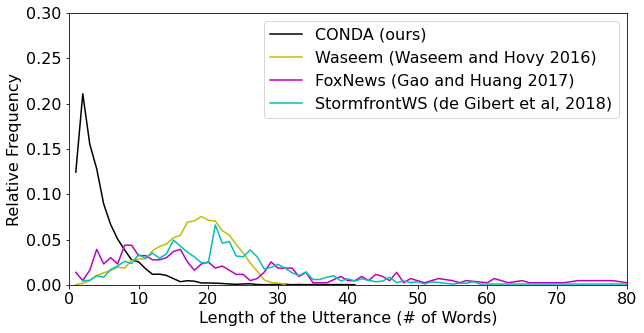}
    \caption{Distribution of toxic utterance length across similar datasets.}
    \label{fig:transfer}
\end{figure}

\paragraph{Comparison with Other Datasets}
In Figure \ref{fig:transfer}, we compare our dataset with other toxicity detection datasets using the metric of relative frequency of toxic utterances of each length. The datasets we compare with are \textbf{1) Waseem} \cite{WaseemH16} which consists of 16.2k tweets binary classified as racism/sexism or other, \textbf{2) FoxNews} \cite{GaoH17} which is 1.5k sentences from Fox News discussion threads classified as hateful/non-hateful, and \textbf{3) StormfrontWS} \cite{gibert2018hate} which is 10.7k conversation sentences from white supremacist website Stormfront classified as hate speech/non-hate speech. For this analysis, we merge classes into toxic/non-toxic as required. As an example of the CONDA dataset, we combine E and I intent classes into a toxic class, and A and O into a non-toxic class.

The distribution in CONDA is different to the other datasets in that the toxic utterances are shorter. This is due to the terseness of in-game chat during playing, with longer utterances occurring in pre-game and post-game discussion. Waseem has a particular distribution due to the character limit in Twitter (140 characters at the time). FoxNews and StormfrontWS are forums which foster the use of longer sentences. 

\section{Baseline Experiment}
To explore the toxicity detection from an NLU perspective, we selected five baseline NLU models and compared their detection performance over our proposed dataset. 

\subsection{Data Preparation}
We split the data into train/validation/test sets in the proportions of 0.6/0.2/0.2, or in samples 26,921/8,974/8,974. The data passed to the models is the tokenised utterances with punctuation removed, and for training the slot and intent labels.

\begin{table*}[t!]
\centering
\scriptsize
\setlength{\tabcolsep}{1.2mm}{
\begin{tabular}{l|l|llll|l|llllll|l}
\toprule
\multicolumn{14}{c}{\textbf{Metrics}} \\
\midrule
\textbf{Model} & \textbf{UCA} &\textbf{U-F1(E)} & \textbf{U-F1(I)} &\textbf{U-F1(A)} &\textbf{U-F1(O)} & \textbf{T-F1} & \textbf{T-F1(T)} &\textbf{T-F1(S)} &\textbf{T-F1(C)} &\textbf{T-F1(D)} & \textbf{T-F1(P)}  & \textbf{T-F1(O)} &\textbf{JSA}  \\
\midrule
\makecell[l]{\textbf{RNN-NLU}\\\cite{liu2016attention}} &0.905 &0.813 &0.720 &0.783 & 0.944& 0.970& 0.931& 0.981 &0.930 &0.718 & 0.991 & 0.987 &0.854  \\
\makecell[l]{\textbf{Slot-gated}\\\cite{goo2018slot}} & 0.894 & 0.806 & 0.694 & 0.773 & 0.938& \textbf{0.991}  & \textbf{0.978} & \textbf{0.992} &\textbf{0.982} & \textbf{0.952} &0.997 & \textbf{0.994} & 0.875 \\
\makecell[l]{\textbf{Inter-BiLSTM}\\\cite{wang2018bimodel}} &0.869 & 0.719 & 0.590 & 0.728 & 0.923& 0.865 & 0.871 & 0.889 & 0.869 &0.788 &0.942 & 0.924 & 0.711 \\
\makecell[l]{\textbf{Capsule NN}\\\cite{zhang2019capsule}} & 0.876 &0.735 & 0.706 & 0.643 & 0.926& \textbf{0.991} & 0.975 & 0.991& \textbf{0.982} & 0.949 & 0.997 &\textbf{0.994} & 0.855 \\
\makecell[l]{\textbf{Joint BERT}\\\cite{castellucci2019multi}} &\textbf{0.921} &\textbf{0.872}& \textbf{0.768} &\textbf{0.800} &\textbf{0.954} &0.989& 0.972  &\textbf{0.992}  & 0.979 & 0.914  &\textbf{0.998} & 0.993 &\textbf{0.895} \\
\bottomrule
\end{tabular}
}
\setlength{\belowcaptionskip}{-10pt}
\caption{Joint intent classification and slot labeling performance on CONDA for the five NLU baseline models. It is measured in the four multi-level metrics including: UCA (Utterance Classification Accuracy); the break-down U-F1 for each intent class - E (Explicit), I (Implicit), A (Action), O (Other); the overall T-F1 and breakdown for each slot class -  T (Toxicity), S (game Slang), C (Character), D (Dota-specific), P (Pronoun), O (Other); and JSA (Joint Semantic Accuracy).}
\label{tab:baseline_res}
\end{table*}

\subsection{Baseline NLU Models}
The five NLU models are as follows:
\begin{itemize}
  \itemsep0em
  \item \textbf{RNN-NLU} \cite{liu2016attention} is an attention-based bi-directional recurrent neural network model that jointly predicts the current slot and the intent at each time step using shared hidden states and attention.
  \item \textbf{Slot-gated} \cite{goo2018slot} is an attention-based BiLSTM model which builds on separate attended context for slot filling and intent classification while explicitly feeding the intent context into the process of slot filling via a gating mechanism.
  \item \textbf{Inter-BiLSTM} \cite{wang2018bimodel} combines two inter-connected BiLSTMs performing slot filling and intent classification respectively. The information flow between the two tasks occurs by passing the hidden states at each time step from each side to the other to support the decoding process.
  \item \textbf{Capsule NN} \cite{zhang2019capsule} is a capsule-based neural network that explicitly captures the semantic hierarchical relationship among words, slots and intents via a dynamic routing-by-agreement schema.
  \item \textbf{Joint BERT} directly utilizes the merit of pre-trained BERT \cite{devlin2019bert} and non-recursively conducts the joint prediction over the [CLS] token embedding for intent and the sequence of token embeddings for slots.
\end{itemize}

\subsection{Evaluation Metrics}

We propose to use the following four metrics for conducting a multi-aspect evaluation. The first two follow the existing traditional abusive language detection research for utterance level detection evaluation while the others are the metrics used for slot-level prediction evaluation and the joint task in NLU models. 
\begin{enumerate}
\itemsep0em
\item[(1)] \textbf{UCA:} \textbf{U}tterance \textbf{C}lassification \textbf{A}ccuracy measures the sentence-level classification performance based on the ratio of the number of correctly predicted utterance to the total number of utterances.
\item[(2)] \textbf{U-F1:} \textbf{U}tterance \textbf{F1} score calculates the F1 score for each utterance class.
\item[(3)] \textbf{T-F1:} \textbf{T}oken \textbf{F1} score focuses on the prediction performance for slot tokens and calculates an F1 for each class and the token-based micro-averaged F1 score over all classes excluding label O.
\item[(4)] \textbf{JSA:} \textbf{J}oint \textbf{S}emantic \textbf{A}ccuracy measures the overall prediction performance over the semantic hierarchy. An utterance is deemed correctly analysed only if both utterance-level and all the token-level labels including Os are correctly predicted.

\end{enumerate}




\subsection{Implementation details}
Links to the source code are given in Appendix \ref{sec:sourcecode}. For Joint BERT, Slot Gated SLU and Capsule NN, we set the number of epochs as 2, 8 and 60, respectively. For the RNN-NLU model, the global step is 1,200 and bidirectional RNN is used with the attention mechanism. For other hyper-parameters, the configuration for the best model in the official GitHub implementation of the baseline models is used. All the experiments are conducted on 16GB Tesla V100-SXM2 GPU with CUDA 10.1.

\subsection{Baseline results}
The experiment result is provided in Table \ref{tab:baseline_res}. In columns 2 to 6, the metrics associated with utterance labels are shown. The UCA ranges between 0.87 (Inter-BiLSTM) and 0.92 (Joint BERT), and the U-F1 also illustrates a variance of results for each intent class. We observe that class O always achieves the highest F1 score due to its dominance in numbers throughout the dataset. Comparatively, class I presents relatively low F1 scores due to its subtle nature and reliance on understanding context. The variance in U-F1(I) implies potential improvement in implicit toxicity detection. 

Columns 7 to 13 present the metrics related with token labels and show much higher overall performance than utterance labels. This indicates the lexicon-based slot automation gives underlying patterns the model can learn easily. Even so, slot class D always has a lower T-F1 score than other slot classes, indicating game-specific tokens in class D have flexible and variant forms, which increases the difficulty of detection. In the last column, the JSA which jointly handles utterance and token labels is shown. Due to the limitation on utterance level intent classification, it presents comparably low JSA scores, indicating a challenge for improvement. 

Amongst the models, the non-recursive Joint BERT model performs the best due to the rich linguistic information learned in pre-training. Joint BERT has an implicit influence between the intent sub-task and the slot sub-task based on a joint loss, whereas the recursive models Slot-gated and Capsule NN have explicit influence flowing from intent to slot, leading to similar slot performance. These explicit lines of influence from one task to the other have shown to be successful in NLU and could be explored further in the toxicity detection task.

\section{Transfer Experiment}
We compared our dataset with the toxicity detection datasets introduced in Section \ref{sec:datasetstats} in terms of transfer performance over utterance-level binary prediction as toxic or non-toxic. That is, training on one dataset and testing on the others. For simplicity, we solely use the intent classification circuit of the Joint BERT as the prediction model. 

\subsection{Data Preparation}
We combine classes into toxic/non-toxic as explained in Section \ref{sec:datasetstats}. For each dataset, we split into train/test sets in the ratio of 0.9/0.1. The statistics for each dataset are shown in Table \ref{tab:transfer_data}.



\begin{table}[h]
\centering
\small
\setlength{\tabcolsep}{9.6mm}{
\begin{tabular}{lr}
  \toprule
  \textbf{Dataset} & \textbf{Train / Test} \\
   \midrule
  Waseem & 14,581 / 1,621 \\ 
  StormWS & 9,849 / 1,621 \\ 
  FoxNews & 1,373 / 1,095 \\ 
  CONDA (ours) & 40,382 / 4,487 \\ 
  \bottomrule
\end{tabular}}
\setlength{\belowcaptionskip}{-10pt}
\caption{Dataset sample counts for transfer experiment.}
\label{tab:transfer_data}
\end{table}


\subsection{Transfer results}
The transfer performance measured in UCA is given in Table \ref{tab:transfer}. Firstly, we look at the test performance of each dataset trained on the other datasets, that is to compare results in each column. It can be seen from column 4 that CONDA's transferred performance is generally good when trained on each of the other three datasets, ranging from 0.81 to 0.83. This implies that CONDA covers the nature of toxicity that can be generalized from the other toxicity datasets which emphasize hatefulness, sexism and racism. 

Changing the perspective to the rows, we compare the test performance for each dataset on a model on another dataset. StormfrontWS and CONDA perform well on Waseem, picking up the racism components there. However Waseem does not perform well when trained on either of those, suggesting their specific hate speech and game speech respectively is too focused. FoxNews training transfers the weakest results indicating its general news nature is too broad. 

StormfrontWS performs well when trained on CONDA due to shared toxicity characteristics, but the performance of Waseem and FoxNews when tested on a model trained on CONDA is relatively low. We propose that this is due to two aspects of our dataset previously discussed: the specific game related nature of our language, and the shorter utterances in our set compared to the others.

\begin{table}[h]
\small
\setlength{\tabcolsep}{1.2mm}{
\begin{tabular}{ll|cccc}
\toprule
 & & \multicolumn{4}{c}{\textbf{Testing}} \\
 & & Waseem & StormWS & FoxNews & \textbf{CONDA} \\ 
\midrule
\multirow{4}{*}{\rotatebox{90}{\textbf{Training}}} 
 & Waseem & - & 0.8845&	0.7287	&0.8307 \\  
 & StormWS & 0.7118 &-&	0.7379&	0.8329 \\  
 & FoxNews & 0.6931	& 0.8241	& -	&0.8056 \\ 
 & \textbf{CONDA} & 0.6955&	0.8748&	0.6690& -	  \\ 
\bottomrule
\end{tabular}
}
\caption{Transfer Learning Result, UCA (Utterance Classification Accuracy).}
\label{tab:transfer}
\end{table}




\section{Conclusion and Future Work}

In this paper, we propose CONDA, a new dataset with dual-level (token and utterance) annotation for understanding in-game chat and to detect toxicity. Compared to previous studies, we draw on the NLU perspective and use the joint token-utterance aspect for detection of toxicity. Accordingly, we formalise a multi-level evaluation system. Through experiments with joint slot and intent NLU models, we show the promising potential of such models for toxicity detection utilizing the dual-level annotation. We also compare our dataset with other benchmark toxicity datasets in the literature through a transfer experiment. In future work, the automated token labelling can be manually adjusted and the size of the dataset can be expanded.







\section{Ethics/Broader Impact Statement} \label{sec:ethic}
The study follows the ethical policy set out in the ACL code of Ethics\footnote{https://www.aclweb.org/portal/content/acl-code-ethics} and addresses the ethical impact of presenting a new dataset. In addition, it is approved by our Institutional Review Board (project number : 2019/741). As described in the data collection section, our annotated dataset, CONDA, is based on the \textit{Dota 2} game chat where it can be accessed on Kaggle website (See Section \ref{data_collection}). 

For our automated slot labelling, we generated the game toxicity lexicon by taking the supplemental materials released by \citet{MartensSIK15} and \citet{ElSheriefNNVB18} and the list of words banned by Google\footnote{https://github.com/RobertJGabriel/Google-profanity-words}. We then added variants or new toxic words found in the utterances extracted from Kaggle. For intent labelling, all volunteer annotators were recruited from academia and research students. They were informed about toxic behavior in online games before handling the data. Our instructions allowed them to feel free to leave if they were uncomfortable with the content. Due to privacy considerations, we group them by online game experiences and do not take into account annotators’ demographic information. 

The CONDA dataset is intended for toxicity detection in online games by providing both slot and intent labels. With respect to the potential risks, we note that the subjectivity of human annotation would impact on the quality of the dataset. In order to improve the quality of our dataset, we compared the inter-rater agreements between a gamers' group and a non-gamers’ group, and then final annotation of the whole dataset was performed by gamers only.

\setlength{\parskip}{0em}
\bibliographystyle{acl_natbib}
\bibliography{anthology,acl2021}

\newpage
\appendix
\section*{Appendix}
\label{sec:appendix}
\addcontentsline{toc}{section}{Appendix}
\renewcommand{\thesubsection}{\Alph{subsection}}

\subsection{The CONDA datasets}
\label{sec:conda_column}
The CONDA dataset consists of 9 columns as follows:
\begin{itemize}
  \itemsep0em
  \item {\fontfamily{qcr}\selectfont matchId} (numeric): Each match has a unique ID from raw data.
  \item {\fontfamily{qcr}\selectfont conversationId} (numeric): Each conversation has a unique ID generated by this research to provide guidance for human annotation.
  \item {\fontfamily{qcr}\selectfont playerId} (alphanumeric): Individual players have a unique ID from raw data.
  \item {\fontfamily{qcr}\selectfont playerSlot} (numeric): Individual players have a unique number associated with their roles in each match. 
  \item {\fontfamily{qcr}\selectfont chatTime} (numeric): Each utterance has time (in seconds) when it appears in each match. For example, an utterance occurring 10 minutes after starting the game has a chatTime of 600.
  \item {\fontfamily{qcr}\selectfont utterance} (alphanumeric): Original raw data before any data processing (e.g. `retard sf…').
  \item {\fontfamily{qcr}\selectfont slotTokens} (alphanumeric): Tokenised, cleaned, and slot labelled data (e.g. retard (T), sf (C)). 
  \item {\fontfamily{qcr}\selectfont intentClass} (alphabetic): Utterance-level annotated labels - E (Explicit), I (Implicit), A (Action), and O (Other).
  \item {\fontfamily{qcr}\selectfont slotClasses} (alphabetic): Token-level annotated labels - T (Toxicity), C (Character), D (Dota-specific), S (game Slang), P (Pronoun), and O (Other).

\end{itemize}

\subsection{Word clouds}\label{sec:wordcloud}

\begin{figure} [h]    
     \begin{subfigure}{0.49\linewidth}
         \includegraphics[width=\linewidth]{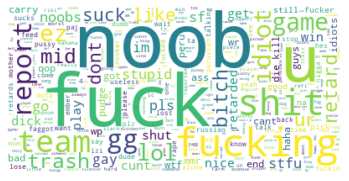}
         \vspace*{-7mm}
         \caption{Class 'Explicit'}
         \label{fig:Ewordmap}
     \end{subfigure}
     \begin{subfigure}{0.49\linewidth}
         \includegraphics[width=\linewidth]{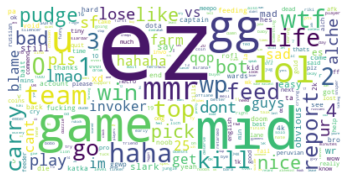}
         \vspace*{-7mm}
         \caption{Class 'Implicit'}
         \label{fig:Iwordmap}
     \end{subfigure}
    \begin{subfigure}{0.49\linewidth}
         \includegraphics[width=\linewidth]{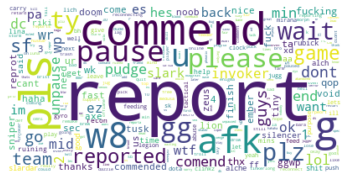}
         \vspace*{-7mm}
         \caption{Class 'Action'}
         \label{fig:Awordmap}
     \end{subfigure}
     \begin{subfigure}{0.49\linewidth}
         \includegraphics[width=\linewidth]{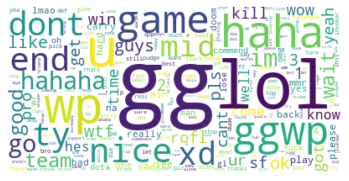}
         \vspace*{-7mm}
         \caption{Class 'Other'}
         \label{fig:Owordmap}
     \end{subfigure}
     \caption{Word clouds for each intent class}
     \label{fig:wordmap}
\end{figure}

The word clouds visualizes the most frequent words associated with each intent class. The top keywords in each class are “noob” in E, “ez” in I, “report” in A, “gg” in O. 

\subsection{Source code}\label{sec:sourcecode}

The source code for the models used to analyse our dataset is available at the following GitHub addresses:

\begin{itemize}
  \itemsep0em
  \item {RNN-NLU}: \\ https://github.com/HadoopIt/rnn-nlu
  \item {Slot-gated}: \\ https://github.com/MiuLab/SlotGated-SLU
  \item {Inter-BiLSTM}: \\ https://github.com/ray075hl/Bi-Model-Intent-And-Slot
  \item {Capsule NN}: \\ https://github.com/czhang99/Capsule-NLU 
  \item {Joint BERT}:  \\ https://github.com/monologg/JointBERT
\end{itemize}

\end{document}